\documentclass[lettersize,journal]{IEEEtran}
\usepackage{amsmath,amsfonts}
\usepackage{array}
\usepackage{textcomp}
\usepackage{stfloats}
\usepackage{url}
\usepackage{verbatim}
\usepackage{graphicx}
\usepackage{cite}
\usepackage{times}
\usepackage{epsfig}
\usepackage{graphicx}
\usepackage{adjustbox}
\usepackage{tabularx} % 用于创建跨栏表格
\usepackage{utfsym}
\usepackage{amsmath}
\usepackage{amssymb}
\usepackage{multirow}
\usepackage{booktabs}

\usepackage{lipsum}
\usepackage{subcaption} % For subfigure environment
\usepackage{multirow}
\usepackage{makecell} % For multiline cells
\usepackage[linesnumbered, ruled]{algorithm2e}
\usepackage[table]{xcolor} % 引入xcolor包以支持表格颜色
\usepackage{hyperref}

\begin{document}

\title{CrossRay3D: Geometry and Distribution Guidance for Efficient Multimodal 3D Detection}

\author{
  Huiming Yang, Wenzhuo Liu, Yicheng Qiao, Lei Yang, Xianzhu Zeng, Li Wang, Zhiwei Li, Zijian Zeng, Zhiying Jiang, Huaping Liu, \textit{Senior Member, IEEE}, and Kunfeng Wang, \textit{Senior Member, IEEE}
  \thanks{Huiming Yang, Wenzhuo Liu, and Yicheng Qiao are equal contributors for this work. 
  (Co-corresponding author: Zhiwei Li and Zhiying Jiang) }
  \thanks{
    Huiming Yang, Zhiwei Li, and Zhiying Jiang are with Beijing University of Chemical Technology, Beijing, 100029, China (e-mail: xuihaipiaoxiang@gmail.com; 2022500066@buct.edu.cn; jiangzy@buct.edu.cn). 

    Wenzhuo Liu and Xianzhu Zeng are with the Division of Energy-Mobility Convergence, Beijing Institute of Technology, Zhuhai, China, 519088 (e-mail: wzliu@bit.edu.cn; xzhuzeng@gmail.com).

    Yicheng Qiao is with the School of Vehicle and Mobility, Tsinghua University, Beijing, 100084 (e-mail: yichengqiao21@gmail.com).

    Lei Yang is with the School of Mechanical and Aerospace Engineering, Nanyang Technological University, Singapore, 639798, Singapore (e-mail: yanglei20@mails.tsinghua.edu.cn).

    Li Wang is with the School of Mechanical Engineering, Beijing Institute of Technology, Beijing 100081, China (e-mail: wangli\_bit@bit.edu.cn).

    Zijian Zeng is with the Institute of Computer Science and Digital Innovation, UCSI University, Kuala Lumpur, 56000, Malaysia (e-mail: 1002266693@ucsiuniversity.edu.my).

    Huaping Liu is with the State Key Laboratory of Intelligent Technology and  Systems and Department of Computer Science and Technology, Tsinghua University, Beijing 100084, China (e-mail: hpliu@tsinghua.edu.cn).   
  }
}

\maketitle
\begin{abstract}
The sparse cross-modality detector offers more advantages than its counterpart, the Bird's-Eye-View (BEV) detector, particularly in terms of adaptability for downstream tasks and computational cost savings. However, existing sparse detectors overlook the quality of token representation, leaving it with a sub-optimal foreground quality and limited performance. In this paper, we identify that the geometric structure preserved and the class distribution are the key to improving the performance of the sparse detector, and propose a Sparse Selector (SS). The core module of SS is Ray-Aware Supervision (RAS), which preserves rich geometric information during the training stage, and Class-Balanced Supervision, which adaptively reweights the salience of class semantics, ensuring that tokens associated with small objects are retained during token sampling. Thereby, outperforming other sparse multi-modal detectors in the representation of tokens. Additionally, we design Ray Positional Encoding (Ray PE) to address the distribution differences between the LiDAR modality and the image. Finally, we integrate the aforementioned module into an end-to-end sparse multi-modality detector, dubbed CrossRay3D. Experiments show that, on the challenging nuScenes benchmark, CrossRay3D achieves state-of-the-art performance with 72.4$\%$ mAP and 74.7$\%$ NDS, while running 1.84$\times$ faster than other leading methods. Moreover, CrossRay3D demonstrates strong robustness even in scenarios where LiDAR or camera data are partially or entirely missing. The code is available on \url{https://github.com/xuehaipiaoxiang/CrossRay3D}.
\end{abstract}

\begin{IEEEkeywords}
Computer Vision, 3D Object Detection, Sparse Detector
\end{IEEEkeywords}

\section{Introduction}
\begin{figure}[htbp]
    \centering
    \begin{subfigure}{0.5\textwidth}
        \captionsetup{skip=2pt}
        \centering
        \includegraphics[width=\linewidth]{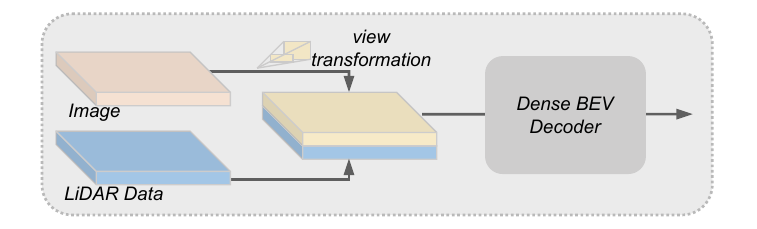}  
        \caption{BEVFusion: Substantial Computational Costs}  
        \label{fig:y1} 
    \end{subfigure}
    \hfill
    \begin{subfigure}{0.5\textwidth}
        \captionsetup{skip=2pt}
        \centering
        \includegraphics[width=\linewidth]{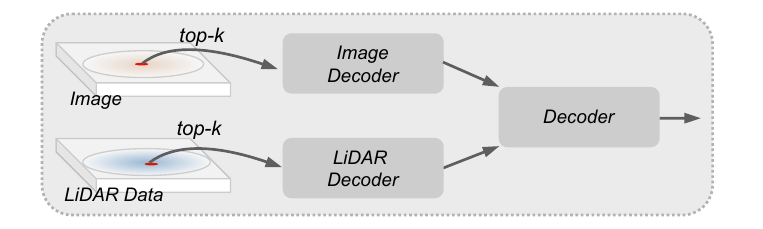} 
        \caption{SparseFusion: Suboptimal Performance} 
        \label{fig:y2} 
    \end{subfigure}
    \hfill
    \begin{subfigure}{0.5\textwidth}
        \captionsetup{skip=2pt}
        \centering
        \includegraphics[width=\linewidth]{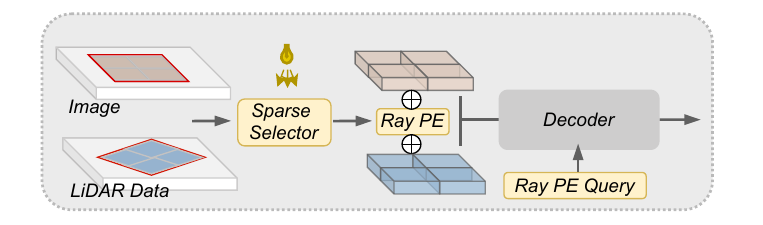}  
        \caption{CrossRay3D (ours)} 
        \label{fig:error}  
    \end{subfigure}
    \caption{Comparison of multimodal 3D object detection methods: (a) BEVFusion, a typical dense detector, is hindered by high computational costs; (b) SparseFusion, a sparse detector, exhibits low performance; (c) In contrast to existing sparse methods, CrossRay3D selects multi-modal instance-level tokens using a sparse selector module, which are then directly fed into the fusion decoder to generate fused predictions, achieving low computational costs and improved performance.}

    \label{fig:start}
\end{figure}

\IEEEPARstart{M}{ultiple} sensor fusion provides significant advantages for 3D detection in improving robustness and safety of autonomous driving system~\cite{arnold2019survey,bai2022transfusion,gong2022multi,li2024mipd}. For instance, LiDAR sensors provide precise geometric information about real scenes~\cite{liang2022bevfusion,yan2023cross}, while images supply rich semantic details about road elements~\cite{cui2024textnerf,gong2023sifdrivenet,liu2024glmdrivenet,tan2025graph,liu2024fmdnet,bi2025vm,tan2025samoccnet,gan2024segmentation}. However, for real-world perception, detectors~\cite{liu2023bevfusion, chen2022deformable, yin2024fusion} that rely on highly structured BEV (Fig.~\ref{fig:start} (a)) limit the adaptability of multi-modality methods to downstream tasks and introduce additional costs in computing background information that is not related to the task of 3D object detection.

To address these challenges, some researchers have started to explore more efficient sparse representations for multi-modality detectors~\cite{bai2022transfusion, yan2023cross, li2022unifying}. For example, SparseFusion~\cite{xie2023sparsefusion} employs a token sampling strategy to mitigate the influence of noisy backgrounds while simultaneously reducing computational overhead (Fig.~\ref{fig:start}~(b)). However, token sampling that relies on class semantic information by a simple top-k operation is suboptimal. Firstly, directly relying on class semantics can lead to missing object boundaries, which are crucial for the decoder to recognize the structure and depth of objects~\cite{zhu2020deformable, zhang2022dino,chen2025effects, meng2021conditional}. Besides, a simple top-k sampling based solely on class salience may harm the recall of all objects. In other words, the operation tends to neglect tokens that correspond to small-sized objects during sampling. That is, the quality of token sampling has yet to be adequately addressed.

In this paper, we further explore the key to improving the quality of sampled tokens. Motivated by rays from the optical center to objects, which naturally reflect the full structure of the objects, we propose that the ray passing through a pixel and hitting an object in 3D scenes can serve as object-structure-oriented supervision to generate high-structure foreground tokens. Additionally, we observe that the class distribution of objects can serve as guidance for learning all objects within the entire scene. As a result, more emphasis is placed on hard examples, while less attention is given to tokens related to easily learned objects during token sampling.

To this end, we propose Sparse Selector (SS) to achieve high-quality token sampling from both geometric and class-balanced perspectives. Initially, the tokens from the image encoder and LiDAR encoder are supervised by Ray-Aware Supervision to predict the salience of each token, which enforces the tokens to generate geometric features related to 3D objects. Then, Class-Balanced Supervision (CBS) loss is employed to reweight the salience of the tokens, which utilizes the distribution of ground truth categories to adaptively scale the weight of tokens related to objects with different scales. We achieve these steps through several convolutional layers with negligible computational cost. Subsequently, the sampled tokens from both LiDAR and camera data are combined with Ray positional encoding (Ray PE) to mitigate the distribution discrepancy from different modalities. Similarly, for directly complementary feature aggregation, we incorporate Ray PE to generate the initial query. Finally, we integrate the aforementioned module in an end-to-end manner and propose CrossRay3D (Fig.~\ref{fig:start} (c)), a sparse multi-modality detector. On the challenging nuScenes benchmark~\cite{caesar2020nuscenes}, our base model achieves 72.4\% mAP, while being 2x faster than the state-of-the-art model \cite{yan2023cross} on a single A40 GPU. To summarize, our contributions are:

\begin{itemize}
    \item We propose Sparse Selector for joint image and LiDAR data token sampling, considering both geometric structure information and class balance, which can function as a plug-and-play module.
    \item The design of RAS and CBS leverages the shape and distribution of 3D objects to generate high-quality geometric and class-balanced tokens, achieving negligible computational cost and significant performance improvements.
    \item We introduce Ray PE to address the distribution discrepancy in directly complementary feature aggregation between image and LiDAR data.
    \item  Extensive experiments are conducted on the nuScenes Dataset. CrossRay3D achieves 72.4\% mAP on the competitive nuScenes benchmark with fewer computational costs and faster inference speed than the state-of-the-art model \cite{yan2023cross}.

\end{itemize}

% \begin{figure}[htbp]
%     \centering
%     \includegraphics[width=\linewidth]{fig/semantic.drawio.png}
%     \caption{Statistics of semantic information processed by sigmoid after the image backbone ResNet-50 \cite{he2016deep}.}
%     \label{fig:semantic}
% \end{figure}

\section{Related Works}
\subsection{LiDAR-based 3D Object Detection}\label{section:B}
LiDAR-based detectors leverage the geometric information provided by point clouds for precise 3D object localization. For outdoor scene detection, existing methods adopt various strategies to process point clouds. Point-based methods \cite{qi2017pointnet, qi2017pointnet++, fan2022fully} directly utilize raw point data to generate 3D predictions, while others transform the unstructured point clouds into regularized voxel \cite{zhou2018voxelnet} or pillar \cite{lang2019pointpillars} formats, enabling feature extraction in the Bird’s Eye View (BEV) plane using standard 2D or 3D backbones. Mainstream LiDAR approaches employ center-based detection heads \cite{yin2021center} or anchor-based methods \cite{zhou2018voxelnet} to predict object categories and regress 3D locations. To mitigate the computational burden of processing LiDAR data, recent studies \cite{chen2023voxelnext, Chen_2022_CVPR, zhang2023fully} leverage sparse \cite{yan2018second} and submanifold \cite{graham2017submanifold} convolutions to improve efficiency. Recent LiDAR works \cite{11015732,10103216,10609805} further advance the field by focusing on real-time detection, data augmentation, and feature completion under data sparsity. Despite the strengths of LiDAR-based methods in precise localization, they still face challenges in capturing rich semantic information within complex 3D scenes.

\subsection{Camera-based 3D Object Detection}\label{section:cam}
Camera-based 3D detection has advanced significantly in recent years. Early works \cite{wang2021fcos3d, yin2021center} focused on monocular cameras, adapting existing 2D detectors \cite{tian2020fcos} by adding extra attributes such as depth, size, and orientation to extend them to 3D tasks. However, in practical autonomous driving applications, surround-view cameras are more commonly used. BEV, as a unified coordinate system, offers substantial advantages in integrating information from multiple camera views. LSS \cite{philion2020lift} has gained increasing attention by mapping surround-view cameras to the BEV space through depth estimation, while subsequent works \cite{huang2021bevdet, li2023bevdepth} further explored lifting image features into a 3D frustum meshgrid by predicting depth distributions. Inspired by DETR \cite{carion2020end}, DETR3D \cite{wang2022detr3d} interprets queries as 3D reference points and projects them into the surround-view images for feature interaction. Similarly, PETR \cite{liu2022petr} implicitly incorporates positional encodings into the image, enabling direct query–feature interaction for parallel computation. Recently, some works \cite{li2022bevformer, Wang_2023_ICCV,huang2024mfe, shi2023bssnet, jiang2024far3d, zhang2023oblique} have also explored temporal modeling in camera-based 3D detection to alleviate the challenges of object pose estimation. Recent camera works \cite{10528251,10287865,liu2025mmtl,wang2023path,liu2025umd,liu2025tem} address depth refinement and occlusion handling to enhance BEV-based 3D detection. Despite the rich semantic information captured by camera images, camera-based methods face challenges with occlusion and locating distant objects due to the lack of accurate depth cues and the limitations of perspective.

\begin{figure*}[htbp]
        \centering
	\includegraphics[width=1.0\textwidth]{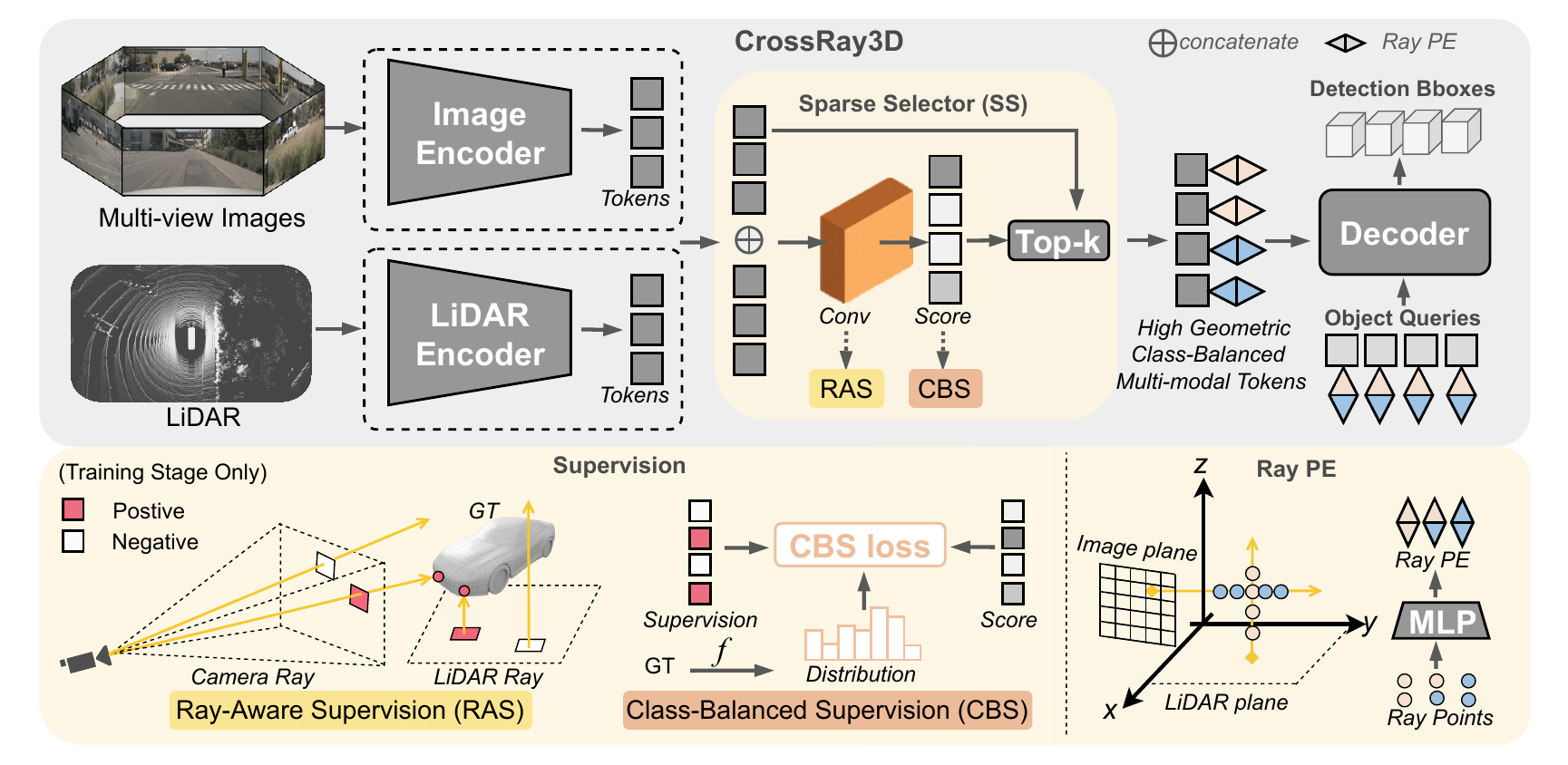}
    \caption{An overview of the architecture of the proposed CrossRay3D is presented. Our CrossRay3D consists of interchangeable LiDAR and image encoders, Sparse Selector (SS), Ray PE, and a Transformer decoder. The supervision for Sparse Selector comes from RAS and CBS, which aim to generate high-quality geometric and class-semantic balanced tokens. With the help of Ray PE, queries interact with sparse multi-modality tokens in an end-to-end manner to predict 3D bounding boxes. Let $f$ represent the function used to analyze the class distribution of GT. }
        \label{fig:main}
\end{figure*}

\subsection{Multi-modal 3D Object Detection}
LiDAR and camera fusion methods have gained significant attention due to the complementary advantages of their modal information. Building on LSS \cite{philion2020lift}, BEVFusion \cite{liu2023bevfusion, liang2022bevfusion} fuses image and LiDAR features in the BEV space. UVTR \cite{li2022unifying}  maps point cloud and image features into a voxel space, using deformable attention \cite{zhu2020deformable} to reduce the computational overhead caused by voxel feature fusion. Inspired by Anchor DETR\cite{wang2022anchor}, FUTR3D \cite{chen2023futr3d} treats queries as 3D reference points and samples features by projecting the reference points onto the corresponding coordinate planes of different modalities. CMT\cite{yan2023cross} introduces positional encoding to both point cloud and image features, enabling direct multimodal feature interaction without explicit feature transformation. These methods effectively leverage the complementary nature of LiDAR and camera data. Recently, lightweight multimodal models have also seen significant advancements, such as TransFusion \cite{bai2022transfusion}, which follows a two-stage pipeline. In this approach, sparse instance-level features are first generated from the LiDAR modality, and then these features are refined by querying image features. Inspired by this, SparseFusion \cite{xie2023sparsefusion} generates sparse instance-level features from both LiDAR and camera inputs using two additional detection heads, which are then fused in the decoder to produce the final results. Overall, due to the inability to directly obtain instance representations, the aforementioned methods rely on multi-stage structures to gradually refine token representations, ultimately generating instance-level features. While these sparse detectors reduce the computational burden of global attention, the multi-stage structure limits their generalizability and introduces additional overhead.

\section{Method}
% We begin by describing the overall architecture of CrossRay3D, followed by an outline of our core contributions. SS are consist  (a) a multi-modality tokens selector dubed SS, which enables high quality and class balanced token generation; (b) the use of 3D ground truth distribution as supervision to adjust semantic weights across scales, ensuring balanced foreground sampling; (c) the introduction of Ray PE to measure the distances between multimodal tokens in 3D, helping mitigate differences in data distribution across modalities.
\subsection{Network Overview}
The overall architecture is illustrated in Fig.\ref{fig:main}. First, multi-view images and LiDAR data are processed by two independent backbones to extract their feature tokens. In the Sparse Selector (SS) module (Sec.\ref{sec:ss}), Ray-Aware Supervision (RAS) is used to guide the model in learning object geometric structures and predicting salience scores for multi-modal tokens. Class-balanced Supervision (CBS) is then applied to reweight the scores, resulting in class-balanced token sampling. Finally, Ray PE (Sec.\ref{sec:Ray PE}) mitigates the distribution discrepancy between LiDAR data and images, and all modalities are jointly learned within a Transformer decoder, outperforming current multi-modality approaches in both efficiency and effectiveness.

\subsection{Sparse Selector for Multi-Modality}\label{sec:ss}
Sparse detectors~\cite{jiang2024far3d,wang2023focal,wang2023exploring,roh2021sparse} have observed that objects of interest occupy only a small fraction of the 3D space. Reducing background tokens strengthens spatial priors and reduces the computational cost of attention, thereby accelerating inference. The key challenge for sparse detectors is to generate high-quality foreground tokens while maintaining a consistent data distribution and preserving essential information. First, we apply the ray–box intersection principle to jointly supervise the geometry of image tokens and point cloud tokens. Tokens are labeled as positive when the rays originating from their corresponding positions intersect with the ground-truth (GT) boxes within the scene.

Specifically, we first construct rays for each pixel in the camera plane and for each cell in the BEV plane based on the camera model and vertically upward directions. The pixel is marked as positive if its corresponding ray intersects a GT box. $\mathbf{F}$ represents the image features from the camera or the point cloud features from the LiDAR. Formally, rays are denoted as $\mathbf{R}^{(i,j)}$ for each spatial location $(i,j)$ on the feature map $\mathbf{F}$, where $(i,j)$ refers to the spatial indices of $\mathbf{F}$. Then, the corresponding feature pixel $\mathbf{F}^{(i,j)}$ is designated as positive and denoted by $\mathbf{G}^{(i,j)}_{F}$. The calculation is given by:

\begin{equation}
\label{ray_hit}
\mathbf{G}^{(i,j)}_{F} =
\begin{cases}
1, & \operatorname{Intersect}\bigl(\mathbf{R}^{(i,j)}, \mathbf{G}\bigr) \\[1ex]
0, & \mathrm{otherwise}.
\end{cases}
\end{equation}

\noindent\textbf{RAS for Image.}
For each surround camera $k \in \{1,\dots,K\}$, every pixel $\mathbf{F}_k^{(i,j)}$ corresponds to a ray originating from the optical center $\mathbf{O}_k$. Then the direction of the ray $\tilde{\mathbf{D}}_{k}^{(i,j)}$ can be formulated as:
\begin{equation}
\tilde{\mathbf{D}}_{k}^{(i,j)}
= \mathbf{S}_{k}\,\mathbf{F}_{k}^{(i,j)} - \mathbf{O}_{k},
\end{equation}
where $\mathbf{S}_k$ denotes the downsampling stride on the image feature map $\mathbf{F}_k$. Let $\mathbf{K}_k \in \mathbb{R}^{3 \times 3}$ denote the intrinsic calibration matrix of camera $k$. We then project the ray direction from the camera frame to the LiDAR coordinate system:
\begin{equation}
    \mathbf{D}_{k}^{(i,j)} = \mathbf{T}_{k}\,\mathbf{K}_{k}^{-1}\,\tilde{\mathbf{D}}_{k}^{(i,j)},
\end{equation}
where $\mathbf{D}_{k}^{(i,j)}$ denotes the transformed ray direction, and $\mathbf{T}_{k}$ represents the transformation matrix from the $k$-th camera to the LiDAR coordinate system. Finally, the ray corresponding to pixel $(i,j)$ in camera $k$ is formulated as:
\begin{equation}
    \label{eq:ray}
    \mathbf{R}^{(i,j)}_k = \mathbf{O}_k + t \mathbf{D}^{(i,j)}_k, \quad t \in \mathbb{R}.
\end{equation}
where $t \in \mathbb{R}$ parameterises the distance from the optical center $\mathbf{O}_k$ along the ray direction $\mathbf{D}^{(i,j)}_k$.

\noindent\textbf{RAS for LiDAR Data.}
Intuitively, BEV encodes the complete 3D scene in the LiDAR coordinate system. For LiDAR data, our RAS constructs rays that extend vertically upward in the BEV space to intersect with the GT boxes. Pixels on the LiDAR plane are labeled as positive if their corresponding rays intersect a GT box, enabling straightforward determination of positive supervision pixels in the BEV according to Equation~\eqref{ray_hit}.

Under RAS supervision, token sampling is performed through multiple convolutional operations. This design makes our approach both computationally efficient and effective, as substantiated by the visualization results presented in Fig.~\ref{fig:heatmapCompare}.

\noindent\textbf{CBS for Distribution Supervision.}
% \subsection{}
\label{sec:dis}
Our objective is to achieve efficient sparse 3D object detection through class-balanced foreground sampling. Inspired by focal loss~\cite{lin2017focal}, we design a distribution-aware supervision scheme, termed Class-Balanced Sampling loss (CBS), where a weight factor is introduced to enhance the model’s sensitivity to foreground tokens while ensuring class balance. Specifically, we first compute the per-class distribution of 3D ground-truth instances in each scene. Based on this distribution, a dynamic top-$k$ strategy is applied to select semantically salient tokens that match the class ratios, which serve as foreground tokens. We further modulate the contribution of all tokens to the classification loss according to semantic weights $\mathbf{W}_n$, where $n$ indexes the tokens, enabling class-balanced foreground perception. Tokens that are not selected as foreground are regarded as background tokens, and their semantic scores are smoothly down-weighted using a sigmoid function to suppress gradients. This design reduces background clutter, provides stronger spatial priors, and improves the efficiency and accuracy of sparse 3D object detection.

\begin{equation}
\mathbf{W}_n =
\begin{cases}
\lambda, & n \in \mathcal{D}, \\[3pt]
\sigma(\max_{c}\hat{y}_{n,c}), & n \notin \mathcal{D},
\end{cases}
\end{equation}
\noindent
where $\mathcal{D}$ denotes the set of tokens selected according to the class distribution, and $\sigma(\cdot)$ is the sigmoid function. The hyperparameter $\lambda \geq 1$ controls the weight assigned to selected tokens. For an ablation study of $\lambda$, see Table~\ref{tab:topk}. We utilize the logits of the $n$-th token to represent the probability distribution over each class \( c \in \{1, 2, \dots, C\} \), denoted as \( \hat{y}_{n,c} \). The loss for the $n$-th sample in the minibatch is computed as:

\begin{equation}
  l_n = -\mathbf{W}_{n} \log \left( \frac{\exp(\hat{y}_{n,y_n})}{\sum_{c=1}^{C} \exp(\hat{y}_{n,c})} \right).
\end{equation}

\begin{algorithm}[!htbp]
\caption{Class-Balanced Supervision (CBS)}
\label{distribution_injection}
\SetKwFunction{FMain}{DistributionInjection}
\SetKwProg{Fn}{Function}{:}{}
\KwIn{$\hat{\mathbf{y}}_n$: Class logits for each token\;}
\KwIn{$\mathbf{T}_n$: Token set for each sample\;}
\KwIn{$\mathbf{W}_n$: Weight for each token\;}
\KwOut{$L$: The CBS loss\;}

\textcolor[HTML]{006400}{\/// Generate class distribution from GT} \\
$\mathbf{P}_n, \mathbf{I}_n \leftarrow \max(\hat{\mathbf{y}}_n)$ \;
\textcolor[HTML]{006400}{\/// Initialize an empty dictionary bag}\;
\For{$i = 1$ \textbf{to} $C$}{
    $\mathrm{cls\_num} \leftarrow \mathrm{sum}(\mathbf{I}_n = i)$\;
    $\mathrm{bag}[i] \leftarrow \mathrm{cls\_num}$\;
}
$\mathbf{W}_n \leftarrow \sigma(\mathbf{P}_n)$\;

\textcolor[HTML]{006400}{\/// Apply Class-Balanced Supervision} \\

\For{$i = 1$ \textbf{to} $C$}{
    $\mathrm{cls\_num} \leftarrow \mathrm{bag}[i]$\;
    \If{$\mathrm{cls\_num} > 0$}{
        $\mathrm{topk\_index} \leftarrow \mathrm{topk}(\mathrm{cls\_num}, \mathbf{I}_n)$\;
        $\mathbf{W}[\mathrm{topk\_index}] \leftarrow \mathbf{W}[\mathrm{topk\_index}] \cdot \lambda$\;
    }
}

$L \leftarrow \mathrm{mean}\bigl(\mathbf{W}_n \cdot \mathrm{CE}(\hat{\mathbf{y}}_n, \mathbf{y}_n)\bigr)$\;
\KwRet{$L$}
\end{algorithm}

We utilize Algorithm~\ref{distribution_injection} to illustrate the distribution of the statistical GT and use the CBS loss to adjust the semantic weights of instances at different scales based on distribution information as supervision during the training process.

% 3D position information into
\subsection{Ray Positional Encoding}\label{sec:Ray PE}
The sampled multimodal sparse tokens exhibit significant variation in data distribution, which creates challenges for direct query interactions. Moreover, to better exploit the spatial prior provided by the Sparse Selector, each query should jointly encode the positional relationships across modalities. Therefore, we propose Ray Positional Encoding (Ray PE), designed to map both the camera and BEV positional encodings into a unified 3D space for simultaneous foreground feature aggregation. Specifically, we sample 3D anchor points along rays originating from both the camera and LiDAR fields. Subsequently, position encodings generated from these 3D anchor points are utilized to measure the distances of foreground multimodal tokens within this 3D space. For query generation, we treat each query as the intersection of a camera ray and a LiDAR ray, and sample 3D anchor points along these rays to construct the corresponding position encodings. Furthermore, by incorporating the 3D anchor points from both the LiDAR and camera into the queries, we enable direct interaction between the queries and multimodal features through global attention.

For details, given a fixed value \(d\), we sample \(d\) anchor points along the rays in the camera field, from near to far. In the LiDAR plane, \(d\) anchor points are sampled from bottom to top. The sampled anchor points are ordered by distance, and as the rays intersect, the distance between the two modalities is implicitly measured. Besides, a mapping module $\mathbf{MLP}$, implemented as a two-layer feed-forward network, is used to project the 3D anchor points $\mathbf{P}^{(i,j)}$ into a shared latent space positional encoding $\mathbf{PE}^{(i,j)}$. By incorporating position embeddings along with multimodal foreground tokens and queries, the Ray PE helps mitigate distributional differences between modalities, enabling queries to interact with data from both modalities simultaneously.

\noindent\textbf{Ray PE for Image.}
Since each pixel \( \mathbf{F}^{(i,j)} \) in the feature map corresponds to a ray \( \mathbf{R}^{(i,j)} \) as described in Equation~\ref{eq:ray}, the position encoding for sparse image tokens can be constructed based on the ray. Specifically, we sample \( d \) points along the ray passing through each pixel. Then, the feature mapping module MLP processes these anchor points, and the position encoding for each pixel is computed as follows:

\begin{equation}
    \label{eq:i}
\mathbf{{PE}}^{(i,j)} = \mathbf{MLP}({\mathbf{P}}^{(i,j)}_d),
\end{equation}
where $\mathbf{MLP}$ takes as input the concatenated features of all sampled points $d$ along the ray of pixel $(i,j)$.

\noindent\textbf{Ray PE for LiDAR.}
Similarly to the image modality, we assign the same position encoding to all points $(i,j)$. We sample $d$ anchor points along the Z axis as a ray, represented by vertical vectors. The corresponding Ray PE in the LiDAR plane feature map is then computed as:

\begin{equation}
    \label{eq:l}
   \mathbf{PE}^{(i, j)} = \mathbf{MLP}(\mathbf{S}\mathbf{P}^{(i,j)}_d),
\end{equation}
where $(i, j)$ represents the size of each BEV feature grid. To simplify, we sample only one point along the height axis, which is different from the 2D coordinate encoding.

\noindent\textbf{Ray PE for Query.}
Different from CMT~\cite{yan2023cross}, which encodes both LiDAR sinusoidal position encoding and camera cone, we directly see the query as two intersecting rays, one from the LiDAR field and the other from the Camera field. The sampled points are then obtained according to Equation~\ref{eq:i} and Equation~\ref{eq:l}. Finally, the feature mapping module is used to map positions into positional embeddings, which are then used to construct queries for unified cross-modal querying.

\subsection{Decoder and Loss}
Following DETR \cite{carion2020end}, we use $L$ original transformer decoder layers to construct our decoder. With the help of Ray PE and the shared latent space, the queries interact directly with multimodal sparse tokens, thereby accelerating the model's computation. After this interaction, two feed-forward networks (FFNs) are applied to the updated queries to predict 3D bounding boxes and object classes.  
The prediction process for each decoder layer can be expressed as follows:
\begin{equation}
    \hat{b}_l = \Phi^{reg}({Q}_l),\quad  \hat{c}_l = \Phi^{cls}({Q}_l),
\end{equation}
Where $\Phi^{reg}$ and $\Phi^{cls}$ represent the feed-forward networks (FFNs) for regression and classification, respectively. $Q_l$ denotes the updated object queries from the $l$-th decoder layer.

Several additional convolutional layers are used to predict foreground scores $\hat{G}$ for the point cloud and image tokens that are highly relevant to the instance. Similar to DETR-based detectors, the CBS loss $\mathcal{L}_{t}$ is obtained as:
\begin{equation}
\mathcal{L}_{t} = 
\mathcal{L}_{t}^{\text{image}}(\hat{{G}}^L, {G}^L_F,B) + 
\mathcal{L}_{t}^{\text{pc}}(\hat{{G}}^C, {G}^C_F,B),
\end{equation}
Here, \( \hat{G}^L \) and \( \hat{G}^C \) represent the salience scores for LiDAR and camera, respectively, while \( {G}^L_F \) and \( {G}^C_F \) denote the supervision from SS.
Additionally, \( {B} \) denotes the statistical distribution supervision derived from the statistical bag, as described in Algorithm~\ref{distribution_injection}. Finally, all modules in our network are optimized in an end-to-end manner. The object classification loss is computed using the focal loss, and the 3D bounding box regression loss is computed using the L1 loss. The overall loss of the framework is defined as:
\begin{equation}
\mathcal{L} = 
\omega_1\mathcal{L}_{t} + 
\mathcal{L}_{\text{cls}}(c, \hat{c}) + 
 \mathcal{L}_{\text{reg}}(b, \hat{b}),
\end{equation}
where $\omega_1$ is a hyperparameter used to balance the CBS loss with box regression and class prediction. We empirically set $\omega_1$ to 1.5.

\begin{figure*}[htbp]
    \centering
    \includegraphics[width=\textwidth]{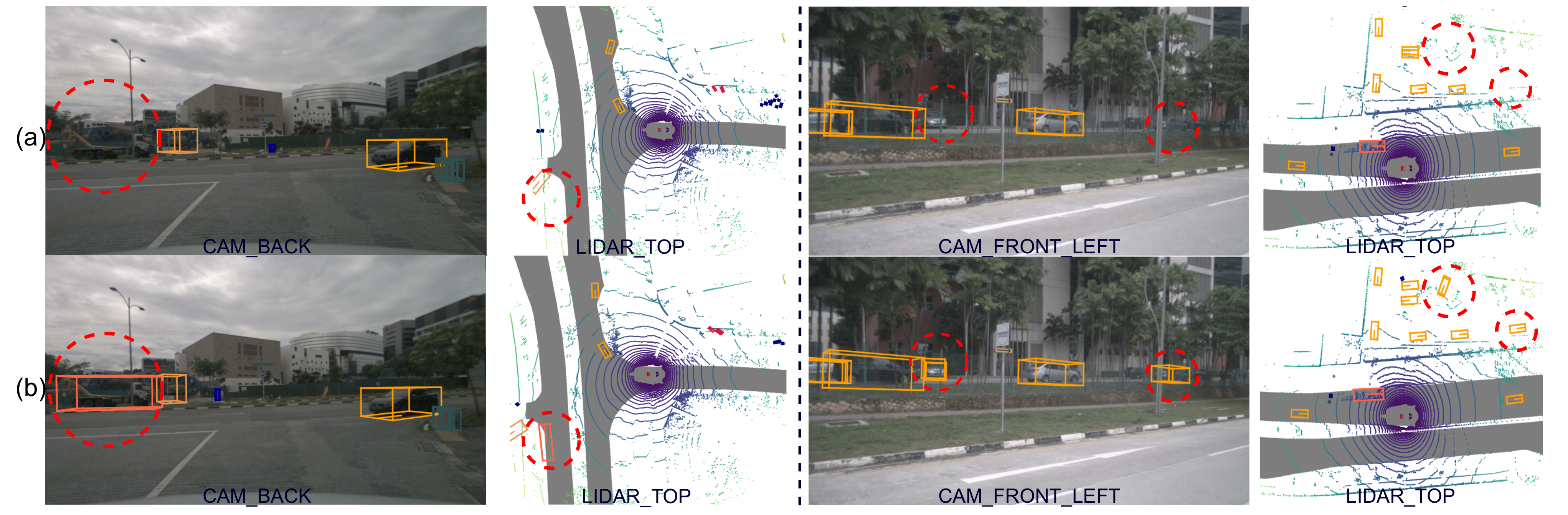}
    \caption{Comparison of baseline SparseFusion (a) and CrossRay3D (b) on the nuScenes validation set. Hard cases, i.e., occlusions and long-distance small-scale instances, are marked with \textcolor{red}{red} circles.}
    \label{fig:sur}
\end{figure*}

\section{Experiments}
\subsection{Datasets and Metrics}
We evaluate our method on the nuScenes dataset \cite{caesar2020nuscenes}, a large-scale, multi-modal benchmark designed for autonomous driving research. NuScenes is highly challenging, comprising data collected from 6 cameras, 1 LiDAR, and 5 radars. The dataset contains 1,000 scenes, which are divided into training, validation, and test sets with 700, 150, and 150 scenes, respectively. Each sequence includes approximately 40 frames of annotated LiDAR point cloud data, accompanied by six calibrated camera images providing a 360° field of view. 

The Argoverse 2 dataset \cite{wilson2023argoverse} comprises 1000 unique scenes, each 15 seconds long, annotated at a rate of 10 Hz. The scenes are divided into 700 for training, 150 for validation, and 150 for testing. The evaluation encompasses 26 categories within a 150-meter range, focusing on long-range perception tasks.

\noindent\textbf{Cameras.} Each scene includes 20 seconds of video captured at 12 FPS. 3D bounding box annotations are provided every 0.5 seconds. For our experiments, we utilize these key frames, with each frame containing images from six cameras.

\noindent\textbf{LiDAR.} NuScenes provides data from a 32-beam LiDAR sensor operating at 20 FPS. Key frames are annotated at the same 0.5-second intervals as the camera data. Following common practice, we aggregate LiDAR points from the previous 9 frames and transform them into the current frame for training and evaluation.

\noindent\textbf{Metrics.} We adopt the official nuScenes metrics for evaluation. Specifically, we report the nuScenes Detection Score (NDS), mean Average Precision (mAP), mean Average Translation Error (mATE), mean Average Scale Error (mASE), mean Average Orientation Error (mAOE), mean Average Velocity Error (mAVE), and mean Average Attribute Error (mAAE). Composite Detection Score (CDS), which integrates three other true positive
 metrics: ATE, ASE, and AOE.

\subsection{Implementation Details}
We use ResNet50 \cite{he2016deep} as the image backbone, with weights loaded from a checkpoint trained on ImageNet \cite{deng2009imagenet}, to extract 2D image features. The C5 feature map is upsampled and fused with the C4 feature map to produce the P4 feature map. For the point cloud backbone, we employ a pure 3D sparse backbone \cite{yan2018second,graham2017submanifold} to extract point-cloud features, initialized with weights from VoxelNeXt. The point cloud region is set to $[-54.0m, 54.0m]$ for the X and Y axes, and $[-5.0m, 3.0m]$ for the Z axis. Six decoder layers are utilized in the vanilla DETR decoder. A voxel size of $[0.1m, 0.1m, 0.2m]$ and an image size of $800 \times 320$ are adopted as the default settings in our experiments. Our model is trained with a batch size of 12 on 2 A40 GPUs over 20 epochs using CBGS \cite{zhu2019class}. We adopt the AdamW \cite{loshchilov2017decoupled} optimizer with an initial learning rate of $1.0 \times 10^{-4}$ and follow the cyclic learning rate policy \cite{smith2017cyclical}. GT sample augmentation is applied during the first 15 epochs and disabled for the remaining epochs.  For fast convergence, we adopt the point-based query denoising strategy from CMT, which introduces noisy anchor points by applying center shifting based on the box size.

\subsection{State-of-the-Art Comparison}
We compare the proposed framework with existing state-of-the-art methods on the validation and test sets of nuScenes~\cite{caesar2020nuscenes}, as well as the large-scale Argoverse 2 dataset. For inference speed comparison, we follow the settings of IS-Fusion~\cite{yin2024fusion}, using a batch size of 1 and FP32 precision on a single RTX 3090 GPU.

\noindent\textbf{nuScenes Test Set.} 
We evaluate CrossRay3D against both sparse and dense detectors, including BEVFusion~\cite{liu2023bevfusion}, TransFusion~\cite{bai2022transfusion}, and CMT~\cite{yan2023cross}. As shown in Tab.~\ref{tab:performance_comparison}, compared with other cross-modality methods, our base model achieves 74.0$\%$ NDS, 71.8$\%$ mAP, and 7.0 FPS, which is 1.84x faster than sparse detector CMT. When using the large-base configuration with a $1600 \times 900$ image resolution, mAP and NDS further improve by 0.7$\%$ and 0.6$\%$, respectively, reaching state-of-the-art performance. In addition, we also conduct single-modality experiments. The LiDAR-only baseline achieves 71.4$\%$ NDS, delivering near state-of-the-art results among all existing LiDAR-only methods. Similarly, the camera-only baseline surpasses mainstream image-based detectors without temporal information in terms of both accuracy and speed.
\noindent\textbf{nuScenes Validation Set.} 
For further fair comparison, we also compare the performance with other SoTA methods on the nuScenes val set (see Tab.~\ref{tab:val}). Our base model achieves 72.4\% NDS and 70.0\% mAP. We further demonstrate the superiority of CrossRay3D through qualitative visualizations presented in Fig.~\ref{fig:sur}.

\noindent\textbf{Argoverse 2 Validation Set.} 
To assess the generalization capability of CrossRay3D, we conduct additional experiments on the Argoverse 2 dataset, as displayed in Tab.~\ref{tab:av2}. Our method significantly outperforms previous SoTA methods, achieving 33.1 $\%$ CDS and 42.4 $\%$ mAP, surpassing PolFusion by an absolute 1.5 $\%$ CDS and 1.8 $\%$ mAP. 

\begin{table*}[htbp]
\centering
\caption{Performance comparison on the nuScenes test set. “L” indicates LiDAR-only input, while “C” indicates camera-only input. CrossRay3D-base uses a voxel size of [0.1,\,0.1,\,0.2], whereas the large model adopts a dual-channel backbone with a finer voxel size of [0.075,\,0.075,\,0.2].}
\label{tab:performance_comparison}
\resizebox{\linewidth}{!}{
\begin{tabular}{r@{}l|c|cc|ccccc|c}
\toprule
\multicolumn{2}{c|}{\textbf{Method}} & Modality & NDS$\uparrow$ & mAP$\uparrow$ & mATE$\downarrow$ & mASE$\downarrow$ & mAOE$\downarrow$ & mAVE$\downarrow$ & mAAE$\downarrow$ & FPS$\uparrow$ \\ 
\midrule
FCOS3D\cite{wang2021fcos3d}  & \textcolor[HTML]{708090}{\scriptsize{[ICCV 21]}} & C  & 42.8 & 35.8 & 69.0  & 24.9 & 45.2 & 143.4 & \textbf{12.4} & 15.0 \\
BEVDet\cite{huang2021bevdet}  & \textcolor[HTML]{708090}{\scriptsize{[21]}}   & C  & 48.8 & 42.4 & \textbf{52.4} & \textbf{24.2} & 37.3 & 95.0  & 14.8          & \textbf{16.7} \\
DETR3D\cite{wang2022detr3d}   & \textcolor[HTML]{708090}{\scriptsize{[CoRL 22]}} & C  & 47.9 & 41.2 & 64.1  & 25.5 & 39.4 & 84.5  & 13.3          & 3.7 \\
PETR\cite{liu2022petr}        & \textcolor[HTML]{708090}{\scriptsize{[ECCV 22]}}  & C  & 50.4 & 44.1 & 59.3  & 24.9 & 38.3 & 80.8  & 13.2          & 8.1 \\
\rowcolor{gray!15} 
\multicolumn{2}{c|}{CrossRay3D} & C  & \textbf{51.5} & \textbf{44.9} & 55.4  & 24.3 & \textbf{36.4} & \textbf{79.4} & 13.7          & 10.4 \\ 
\midrule
CenterPoint\cite{yin2021center}   & \textcolor[HTML]{708090}{\scriptsize{[CVPR 21]}}   & L  & 67.3 & 60.3 & 26.2  & 23.9 & 36.1 & 28.8  & 13.6          & 10.4 \\
UVTR\cite{li2022unifying}          & \textcolor[HTML]{708090}{\scriptsize{[NeurIPS 22]}}  & L  & 69.7 & 63.9 & 30.2  & 24.6 & 35.0 & \textbf{20.7} & \textbf{12.3} & - \\
VoxelNeXt\cite{chen2023voxelnext}  & \textcolor[HTML]{708090}{\scriptsize{[CVPR 23]}}   & L  & 70.0 & 64.5 & 26.8  & 23.8 & 37.7 & 21.9  & 12.7          & \textbf{15.5} \\ 
TransFusion-L\cite{bai2022transfusion} & \textcolor[HTML]{708090}{\scriptsize{[CVPR 22]}}   & L  & 70.2 & 65.5 & \textbf{25.6} & 24.0 & 35.1 & 27.8  & 12.9          & 12.5 \\ 
\rowcolor{gray!15} 
\multicolumn{2}{c|}{CrossRay3D} & L  & \textbf{71.4} & \textbf{66.7} & 29.4  & \textbf{23.6} & \textbf{24.6} & 23.3  & 18.7          & 14.8 \\ 
\midrule
% PointPainting\cite{vora2020pointpainting} & \textcolor[HTML]{708090}{\scriptsize{[CVPR 20]}}   & LC & 61.0 & 54.1 & 38.0  & 26.0 & 54.1 & 29.3  & 13.1          & - \\
PointAugmenting\cite{wang2021pointaugmenting} & \textcolor[HTML]{708090}{\scriptsize{[CVPR 21]}}   & LC & 71.1 & 66.8 & 25.3  & 23.5 & 35.4 & 26.6  & 12.3          & - \\
MVP\cite{yin2021multimodal}        & \textcolor[HTML]{708090}{\scriptsize{[NeurIPS 21]}}  & LC & 70.5 & 66.4 & 26.3  & 23.8 & 32.1 & 31.3  & 13.4          & - \\
FusionPainting\cite{xu2021fusionpainting} & \textcolor[HTML]{708090}{\scriptsize{[ITSC 21]}}   & LC & 71.6 & 68.1 & 25.6  & 23.6 & 34.6 & 27.4  & 13.2          & - \\
UVTR\cite{li2022unifying}          & \textcolor[HTML]{708090}{\scriptsize{[NeurIPS 22]}}  & LC & 71.1 & 67.1 & 30.6  & 24.5 & 35.1 & \textbf{22.5} & 12.4          & - \\
TransFusion\cite{bai2022transfusion}   & \textcolor[HTML]{708090}{\scriptsize{[CVPR 22]}}   & LC & 71.7 & 68.9 & 25.9  & 24.3 & 35.9 & 28.8  & 12.7          & 3.2 \\
BEVFusion\cite{liu2023bevfusion}   & \textcolor[HTML]{708090}{\scriptsize{[ICRA 23]}}   & LC & 72.9 & 70.2 & 26.1  & 23.9 & 32.9 & 26.0  & 13.4          & 4.0 \\
ReliFusion\cite{sadeghian2025reliability} & \textcolor[HTML]{708090}{\scriptsize{[25]}}  & LC & 73.2 & 70.6 & -     & -    & -    & -     & -             & - \\ 
BEVFusion\cite{liang2022bevfusion}   & \textcolor[HTML]{708090}{\scriptsize{[NeurIPS 22]}}  & LC & 73.3 & 71.3 & 25.0  & 24.0 & 35.9 & 25.4  & 13.2          & 4.2 \\
DeepInteration\cite{yang2022deepinteraction} & \textcolor[HTML]{708090}{\scriptsize{[NeurIPS 22]}}  & LC & 73.4 & 70.8 & 25.7  & 24.0 & 32.5 & 24.5  & 12.8          & 2.6 \\
SparseFusion\cite{xie2023sparsefusion} & \textcolor[HTML]{708090}{\scriptsize{[ICCV 23]}}  & LC & 73.8 & 72.0 & -     & -    & -    & -     & -             & - \\ 
CMT\cite{yan2023cross}         & \textcolor[HTML]{708090}{\scriptsize{[ICCV 23]}}  & LC & 74.1 & 72.0 & 27.9  & 23.5 & 30.8 & 25.9  & \textbf{11.2} & 3.8 \\ 
% EA-LSS\cite{hu2023ea}         & \textcolor[HTML]{708090}{\scriptsize{[23]}}      & LC & 74.4 & 72.2 & 24.7  & 23.7 & 30.4 & 25.0  & 13.3          & - \\ 
IS-Fusion\cite{yin2024fusion} & \textcolor[HTML]{708090}{\scriptsize{[CVPR 24]}}  & LC & 74.0 & 72.8 & -     & -    & -    & -     & -             & - \\ 
\rowcolor{gray!15} 
\multicolumn{2}{c|}{CrossRay3D-base} & LC & 74.0 & 71.8 & 27.8 & 23.6 & \textbf{29.1} & 25.9 & 12.3 & \textbf{7.0} \\ 
\rowcolor{gray!15} 
\multicolumn{2}{c|}{CrossRay3D-large} & LC & \textbf{74.7} & \textbf{72.4} & \textbf{24.4} & \textbf{23.1} & 29.3 & 25.6 & 11.8 & 5.2 \\ 
\bottomrule
\end{tabular}
}
\end{table*}

\begin{table}[htbp]
\caption{Performance comparison on the nuScenes validation set. “L” indicates LiDAR-only input, while “C” indicates camera-only input. All FPS values are measured with batch size~1 on a single NVIDIA RTX~3090 GPU.}
\label{tab:val}
\begin{center}
\resizebox{\linewidth}{!} {
\begin{tabular}{l|c|cc|c}
\toprule
Method              & Modality & NDS$\uparrow$   & mAP$\uparrow$   & FPS$\uparrow$  \\
\midrule
FUTR3D \cite{chen2023futr3d}          & LC       & 68.0  & 64.2  & 2.3  \\ 
UVTR \cite{li2022unifying}          & LC       & 70.2  & 65.4  & -  \\ 
AutoAlignV2 \cite{chen2207autoalignv2}          & LC       & 71.2  & 67.1  & -  \\ 
TransFusion \cite{bai2022transfusion}        & LC       & 71.3  & 67.5  & 3.2  \\ 
BEVFusion \cite{liu2023bevfusion}          & LC       & 71.4  & 68.5  & 4.0  \\ 
BEVFusion \cite{liang2022bevfusion}          & LC       & 72.1  & 69.6  & 4.2  \\ 
DeepInteration \cite{yang2022deepinteraction}          & LC       & 72.6  & 69.9  & -  \\ 
SparseFusion \cite{xie2023sparsefusion}       & LC       & 72.8  & 70.4  & 5.3  \\ 
CMT \cite{yan2023cross}                & LC       & 72.9  & 70.3  & 3.8  \\
\rowcolor{gray!15} 
CrossRay3D-base   & LC       & 72.4 & 70.0 & \textbf{7.0}    \\ 
\rowcolor{gray!15} 
CrossRay3D-large          & LC       &  \textbf{73.4}    & \textbf{71.0}   & 5.2    \\ 
\bottomrule
\end{tabular}
}
\end{center}
\end{table}

\subsection{Ablation Study}
We perform ablation studies to validate the effectiveness of the proposed components. All experiments are conducted on our base model with a training duration of 20 epochs.

% \begin{figure}[htbp]
%     \centering
%     \includegraphics[width=0.9\linewidth]{fig/ratio.drawio.pdf}
%     \caption{We analyze the mutual influence of tokens from different modalities at different keeping ratios.}
%     \label{fig:heatmapRatio}
% \end{figure}

\begin{figure}[htbp]
    \centering
    \includegraphics[width=0.9\linewidth]{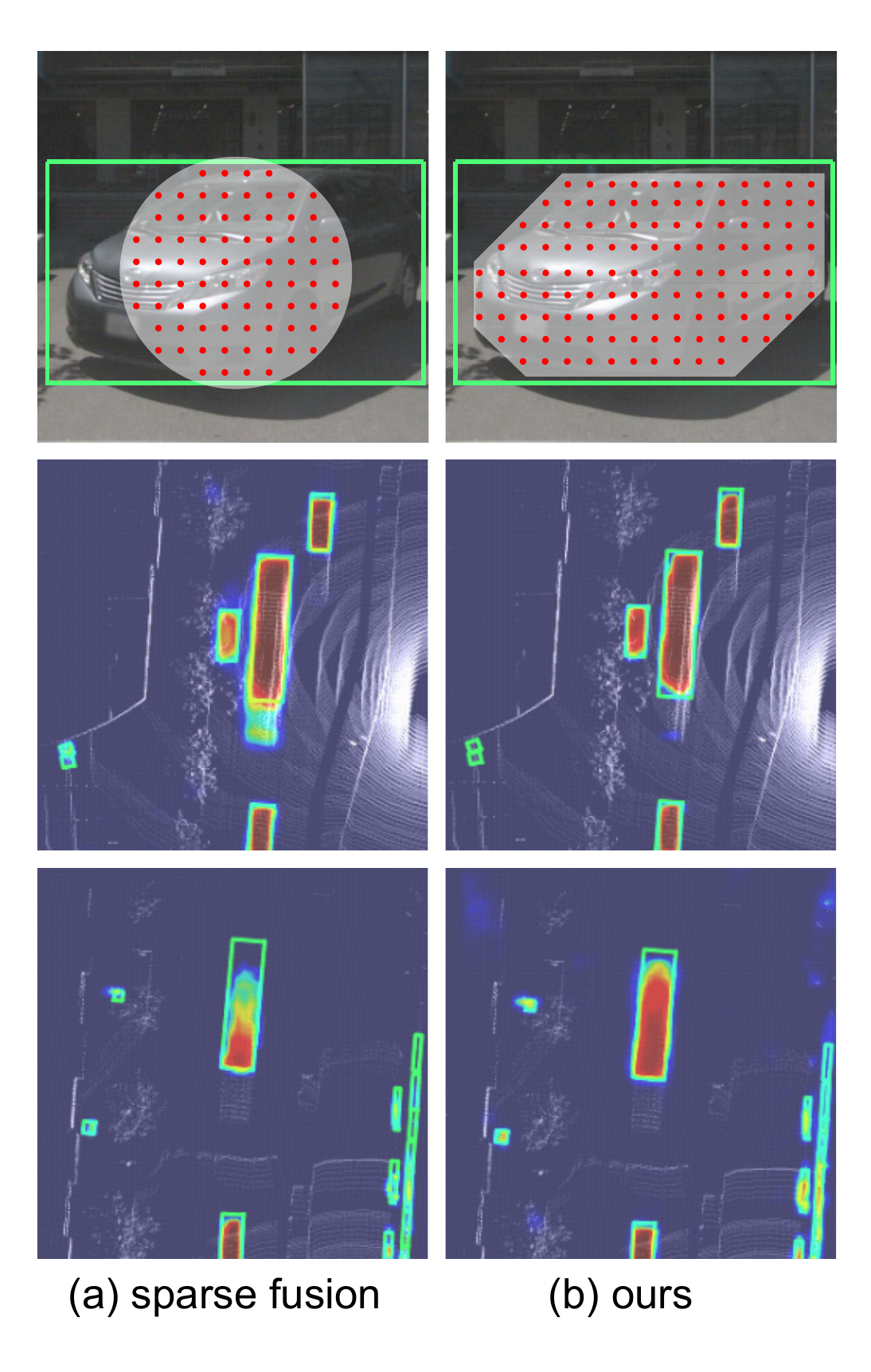}
    \caption{We visualize the supervision from RAS in the camera and the heatmap generated in BEV, comparing it to the baseline model. Note that the \textcolor{green}{green rectangle} represents the original 2D ground truth, and the dots (\textcolor{red}{$\bullet$}) are used to emphasize the supervision.}
    % object-centric~\cite{xie2023sparsefusion}
    \label{fig:heatmapCompare}
\end{figure}

% \begin{table}[htbp]
%     \caption{
%     Ablation study on nuScenes val split.}
%     \label{tab:ablation}
%     \centering
%     \resizebox{\linewidth}{!}{
%     \begin{tabular}{c|cc|c|ccc}
%         \toprule
%         & \multicolumn{2}{c|}{Sparse Selector} & \multirow{2}{*}{Ray PE}   
%         & \multirow{2}{*}{NDS$\uparrow$} & \multirow{2}{*}{mAP$\uparrow$} & \multirow{2}{*}{FPS$\uparrow$} \\
%         & RAS & CBS &  &  &  & \\
%         \midrule
%         (1) &   &   &   & 60.1 & 58.8 & 7.1 \\
%         (2) & \checkmark &   &   & 61.5 & 60.4 & 7.0 \\
%         (3) & \checkmark & \checkmark &   & 61.8 & 60.5 & 7.0 \\
%         (4) & \checkmark & \checkmark & \checkmark & \textbf{72.4} & \textbf{70.0} & 7.0 \\
%         \bottomrule
%     \end{tabular}}
% \end{table}

\begin{table}[htbp]
\caption{
Ablation study with computational overhead analysis on the nuScenes val set. 
The Sparse Selector consists of two modules: RAS and CBS, and is applied only during training. 
The keeping token ratio $\rho$ is fixed to $1.0$ for all experiments in this table.
}
\label{tab:ablation}
\centering
\resizebox{\linewidth}{!}{
\begin{tabular}{c|c|cc|ccc}
\toprule
\multirow{2}{*}{Config} & \multirow{2}{*}{Modules} 
& \multicolumn{2}{c|}{Overhead}
& \multirow{2}{*}{NDS$\uparrow$} & \multirow{2}{*}{mAP$\uparrow$} & \multirow{2}{*}{FPS$\uparrow$} \\
\cmidrule(lr){3-4}
 &  & FLOPs (G) & Mem (GB) &  &  & \\
\midrule
(1) & --                & 504.0   & 18.2 & 60.1 & 58.8 & 7.1 \\
(2) & RAS               & +20.1   & +1.2 & 61.5 & 60.4 & 7.1 \\
(3) & RAS + CBS         & +31.7   & +2.0 & 61.8 & 60.5 & 7.1 \\
(4) & RAS + CBS + Ray PE & +41.2   & +2.1 & \textbf{72.4} & \textbf{70.0} & 7.0 \\
\bottomrule
\end{tabular}}
\end{table}

\begin{figure*}[thbp]
    \centering
    \includegraphics[width=\textwidth]{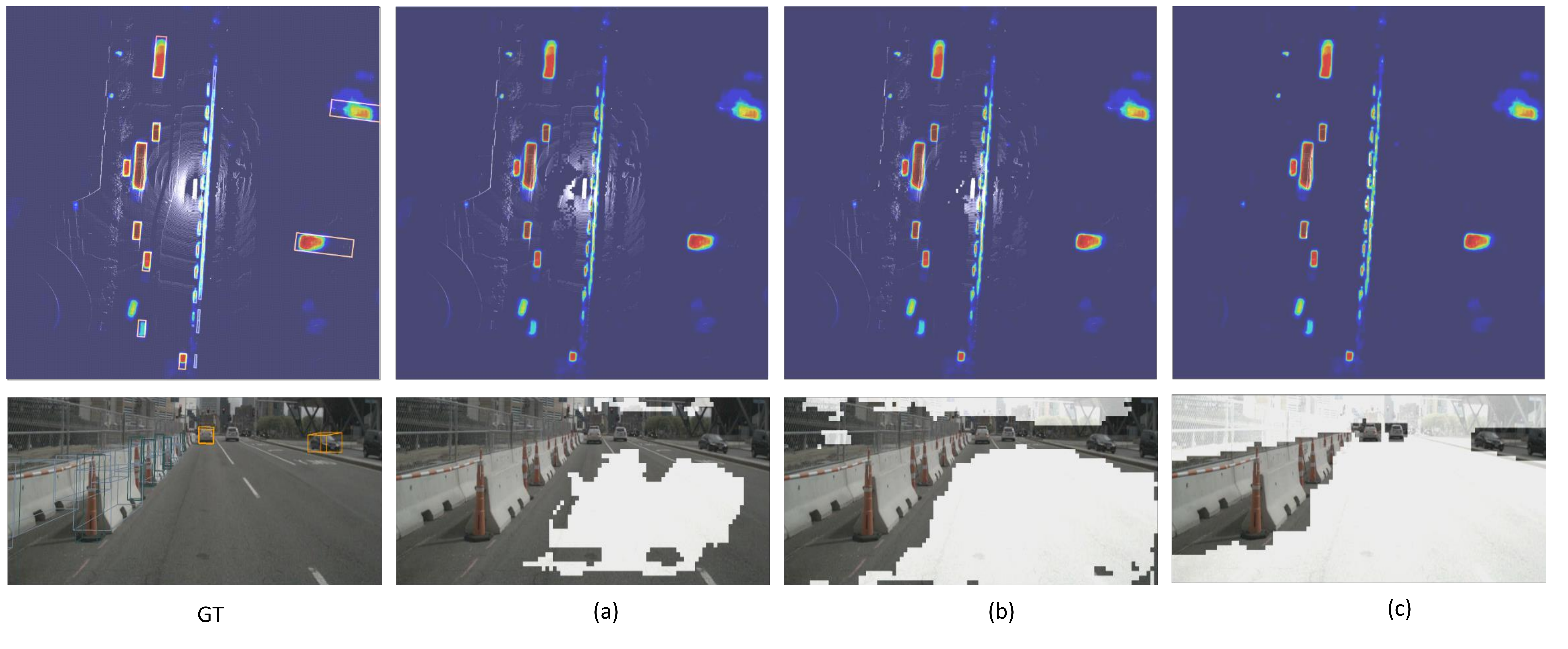}
    \caption{Visualization of sampling locations in point clouds and images with different keeping ratios. Ground truth: (a) 0.75 keeping ratio; (b) 0.5 keeping ratio; and (c) 0.25 keeping ratio. For point clouds, we provide heatmaps based on foreground scores to illustrate the output details of SS in the BEV, where sampled locations are marked in white. For images, redundant tokens are displayed as translucent. This demonstrates that SS can effectively select instance-level representations for both PC and images.}
    \label{fig:sampling}
\end{figure*}

% \begin{table}[htbp]
% \centering
% \caption{Ablation study with different foreground supervision. $\rho$ indicates the keeping ratio for image and point cloud tokens.}
% \label{tab:fore}
% \resizebox{1.0 \linewidth}{!}{
% \begin{tabular}{@{}c|c|cccc@{}}
% \toprule
% $\rho$ &  Strategy &  NDS$\uparrow$ & mAP$\uparrow$ & FLOPs (G)$\downarrow$  & FPS$\uparrow$ \\
% \midrule
% 1.00 & object-centric~\cite{xie2023sparsefusion}  & 72.2 & 69.8  & 537.4 & 6.8\\
%  \rowcolor{gray!15} 
% 1.00 & RAS  & \textbf{72.4} & \textbf{70.0} & 523.2 &7.0\\
% \midrule
% 0.75 & object-centric~\cite{xie2023sparsefusion}  & 70.6 & 68.2  & 492.6 & 6.9\\
%  \rowcolor{gray!15} 
% 0.75 & RAS & \textbf{72.3} & \textbf{70.0} & 480.6 & 7.2 \\
% \midrule
% 0.50 & object-centric~\cite{xie2023sparsefusion}  & 66.7 & 64.3  & 445.6 & 7.1\\
%  \rowcolor{gray!15} 
% 0.50 & RAS & \textbf{70.9} & \textbf{70.0} & 435.1 & 7.3\\
% \midrule
% 0.25 & object-centric~\cite{xie2023sparsefusion}  & 61.1 & 58.4 & 411.2 & 7.4\\
%  \rowcolor{gray!15} 
% 0.25 & RAS & \textbf{70.9} & \textbf{68.1} &  398.4  & 7.6\\

% % \usym{2717} & \usym{2717} &68.8 & 66.4\\
% \bottomrule
% \end{tabular} }
% \end{table} 

\begin{table}[thbp]
\centering
\caption{Ablation studies on the generalization ability of the Sparse Selector, where FLOPs and latency are measured on the same RTX~3090 configuration, and the keeping token ratio $\rho$ is set to $0.5$ to analyze the trade-off between efficiency and accuracy.}
\label{tab:ge}
\resizebox{\linewidth}{!}{
\begin{tabular}{c|c|cccccc}
\toprule
Method & Modality &  NDS$\uparrow$  & mAP$\uparrow$ & FLOPs (G) $\downarrow$ & Params (M)$\downarrow$ &   FPS $\uparrow$ \\ 
\midrule
TransFusion-L~\cite{bai2022transfusion} &  L & 70.2 & 65.5 & 312.7 & 27.9 & 12.5 \\
 \rowcolor{gray!15} 
+Sparse Selector &  L & \textbf{70.6} & \textbf{66.1} & \textbf{234.1} & 28.1 & \textbf{13.0}\\
 \midrule
StreamPETR~\cite{Wang_2023_ICCV} &  C & 54.0 & 43.3 & 410.6 & 57.3 & 27.1 \\
 \rowcolor{gray!15} 
+Sparse Selector &  C & \textbf{54.4} & \textbf{43.6} & \textbf{324.4} & 57.3 & \textbf{27.9}\\
\midrule
CMT~\cite{yan2023cross} &  LC   & 72.9 & 70.3 & 503.9 & 60.5 & 3.8 \\
 \rowcolor{gray!15} 
 +Sparse Selector &  LC   & \textbf{73.6} & \textbf{71.1} & \textbf{398.4} & 60.6 & \textbf{5.2} \\
\bottomrule
\end{tabular} }
\end{table}

\begin{table}[t]
\centering
\caption{
Ablation study of the keeping ratio $\rho$ with deployment-oriented recommendations. 
All results were measured using FP32 precision. The Scenario column indicates typical deployment settings: Edge ($<$4\,GB GPU memory), Mid-range ($<$6\,GB), High-end ($<$6\,GB with higher performance), and Research ($<$8\,GB).}
\label{tab:fore}
\resizebox{\linewidth}{!}{
\begin{tabular}{c|cc|ccc|c}
\toprule
$\rho$ & NDS$\uparrow$ & mAP$\uparrow$ & FLOPs (G)$\downarrow$ & FPS$\uparrow$ & Mem. (GB)$\downarrow$ & Scenario \\
\midrule
0.25 & 70.9 & 68.1 & \textbf{398.4} & \textbf{7.6} & \textbf{3.8} ($<$4GB) & Edge \\
0.50 & 70.9 & \textbf{70.0} & 435.1 & 7.3 & 4.6 ($<$6GB) & Mid-range \\
0.75 & 72.3 & \textbf{70.0} & 480.6 & 7.2 & 5.9 ($<$6GB) & High-end \\
1.00 & \textbf{72.4} & \textbf{70.0} & 523.2 & 7.0 & 7.2 ($<$8GB) & Research \\
\bottomrule
\end{tabular}
}
\end{table}

\begin{table}[htbp]
    \centering
    \caption{Comparison of methods under sensor malfunction conditions.}
    \label{tab:sensor_malfunction}
    \resizebox{\linewidth}{!}{\begin{tabular}{l|c|cc}
        \toprule
        Methods & Sensor Malfunction & NDS$\uparrow$ & mAP$\uparrow$ \\
        \midrule
         BEVFusion\cite{liang2022bevfusion}&  & 54.9 & 45.5 \\
        TransFusion\cite{bai2022transfusion}  & \multirow{2}*{\makecell{Limited LiDAR \\  Field($-90^\circ$, $90^\circ$)}} & 49.2 & 31.1 \\
        SparseFusion\cite{xie2023sparsefusion} &  & 61.2 & 54.3 \\
        CrossRay3D &  & \textbf{61.8} & \textbf{54.9} \\
        \midrule
         BEVFusion\cite{liang2022bevfusion}& & 40.0 & 32.0 \\
        TransFusion\cite{bai2022transfusion}  &  \multirow{2}*{Missing LiDAR}& - & - \\
        SparseFusion\cite{xie2023sparsefusion} &  & - & - \\
        CrossRay3D &  & \textbf{41.3} & \textbf{34.0} \\
        \midrule
        BEVFusion\cite{liang2022bevfusion} & & 70.7 & 65.9 \\
        TransFusion\cite{bai2022transfusion} & \multirow{2}*{\makecell{Missing \\Front Camera}} & 70.1 & 65.3 \\
        SparseFusion\cite{xie2023sparsefusion} &  & \textbf{72.1} & \textbf{69.2} \\
        CrossRay3D &  & 71.3 & 68.5 \\
        \midrule
        BEVFusion\cite{liang2022bevfusion} & & 68.0 & 63.9 \\
        TransFusion\cite{bai2022transfusion} & \multirow{2}*{Missing  Camera}& 70.0 & 65.0 \\
        SparseFusion\cite{xie2023sparsefusion} &  & - & - \\
        CrossRay3D &  & \textbf{70.6} & \textbf{66.5} \\
        \bottomrule
    \end{tabular}}
\end{table}

% \begin{table}[thbp]
% \centering
% \caption{Ablation study with different keeping ratio $\rho$. Note that flash-attention is used during inference stage.}
% \label{tab:keepr}
% \resizebox{1.0\linewidth}{!}{
% \begin{tabular}{c|cc|ccc}
% \toprule
% $\rho$  &  NDS$\uparrow$ & mAP$\uparrow$  & FLOPs (G)$\downarrow$  & FPS$\uparrow$ \\
% \midrule
% 0.25 & 70.9 & 68.1 & 398.4  & 7.6 \\
% 0.50 & 71.9 & 70.0 & 435.1 & 7.3 \\
% 0.75 & 72.3 & 70.0 & 480.6 & 7.2 \\
% 1.00 & 72.4 & 70.0 & 523.2 & 7.0 \\
% % 1.0* & 0.286 & 0.339 & 36.5 & 20.7 \\ 
% \bottomrule
% \end{tabular}
% }
% \end{table}

\subsubsection{Effect of RAS}
As shown in the second row of Tab.~\ref{tab:ablation}, adding RAS yields a notable improvement of $+1.4\%$ in NDS and $+1.6\%$ in mAP. As part of the \textit{Sparse Selector}, RAS (together with CBS) is only applied during training. During training, RAS introduces an additional $20.1$\,GFLOPs and $1.2$\,GB of memory overhead, which is acceptable compared with the overall scale of the network.

To analyze the effectiveness of RAS compared to other foreground supervision methods, we replaced the RAS in the sparse selector with object-centric GT~\cite{xie2023sparsefusion, wang2023focal} supervision. As shown in Tab.~\ref{tab:fore}, with different keeping ratios \(\rho\), RAS demonstrates advantages in both NDS and mAP, attributed to its preservation of the full geometric context. Especially when 25\% of the tokens are retained, RAS still achieves 70.9\% NDS, leading to object-centric supervision by 9.8\% NDS. To further illustrate the effectiveness of our RAS, we present the visualization for the heatmap on BEV as shown in Fig.~\ref{fig:heatmapCompare}. Here, for clearer clarification, we first show the supervision of RAS. In  detector~\cite{xie2023sparsefusion}, the heatmap is ambiguous and out of the range of the ground truth, which leads to sub-optimal performance. On the contrary, with the help of RAS, our method keeps more geometry information and leads to more discriminating heatmaps. Therefore, the decoder can establish more reliable detection results. 

\subsubsection{Effect of CBS}
As shown in the third row of Tab.~\ref{tab:ablation}, when the keeping token ratio $\rho$ is set to $1.0$, the proposed CBS loss yields a $0.3\%$ improvement in NDS. 
To achieve balanced class sampling, an additional $0.8\%$ computational overhead is introduced on top of RAS during training, which remains acceptable relative to the overall computation cost of the network. We further investigate the role of token sampling within the CBS loss. To validate the necessity of CBS for class-balanced supervision, we replace it with standard cross-entropy loss and focal loss~\cite{lin2017focal}. As shown in Tab.~\ref{tab:topk}, the CBS loss outperforms focal loss when the keeping ratio $\rho$ is fixed at 0.5. Notably, for small objects such as \textit{traffic cones}, setting $\lambda = 1$ leads to a $9.7\%$ AP gain over focal loss, demonstrating that CBS effectively achieves class-balanced foreground sampling. In addition, we analyze the impact of the weighting parameter $\lambda$ on NDS and mAP. The results show that performance saturates once $\lambda$ reaches 1.5, indicating that our CBS loss is robust and not sensitive to the exact choice of $\lambda$.

% \begin{figure}[htbp]
%     \centering
%     \includegraphics[width=0.95\linewidth]{fig/fig_distribution.png}
%     \caption{mAP and NDS for different token keeping ratios in images and point clouds. Note that the 'd.s.' denotes the distribution supervision from CBS.}
%     \label{fig:twosubfigTwo}
%     \vspace{-0.2cm}
% \end{figure}

% \begin{table}[htbp]
% \centering
% \caption{Ablation study on the voxel size of the LiDAR backbone. 0.075 represents a voxel size of [0.075, 0.075, 0.2], while 0.100 represents a larger voxel size of [0.1, 0.1, 0.2].}
% \label{tab:image_size}
% \resizebox{1.0\linewidth}{!}{
% \begin{tabular}{c|cccc}
% \toprule
% Voxel size & NDS$\uparrow$ & mAP$\uparrow$ & mATE$\downarrow$& mASE$\downarrow$\\
% \midrule
% 0.075 & 72.6 & 70.3 & 27.6 & 24.0 \\
% 0.100 & 72.7 & 70.0 & 28.1 & 23.8 \\

% \bottomrule
% \end{tabular}
% }
% \end{table}

\subsubsection{Effect of Ray PE} 
As shown in row (4) of Table~\ref{tab:ablation}, introducing positional encoding enables the model to better capture the relative positions of tokens, resulting in a notable mAP improvement of 9.5\%. 
Furthermore, we compare Ray PE with alternative designs, including learnable embeddings and vanilla sinusoidal encodings. 
As reported in Tab.~\ref{tab:raype}, Ray PE proves to be more effective for sparse feature representation in the decoder. 
We argue that sinusoidal positional encoding is suboptimal for additive operations, thereby limiting cross-modal query interaction. 
In contrast, Ray PE is generated directly from position sampling along rays, which facilitates precise measurement of distances across modalities. 
In addition, we investigate the influence of the number of sampled points $d$ along each ray. The results show that when $d$ is set to $16$, the NDS score reaches 72.2\%, highlighting the importance of sufficiently dense ray sampling.

\subsubsection{Analysis of Keeping Ratios $\rho$ on Deployment Cost} 
In this section, we analyze the relationship between the keeping ratio $\rho$, computational resource consumption, and inference speed. The keeping ratio controls the number of salient tokens retained, thereby reducing the computational and memory cost of the decoder. For simplicity, the same $\rho$ is applied to both modalities. As shown in Tab.~\ref{tab:fore}, when the keeping ratio is set to 0.25, peak memory usage is reduced to 3.8\,GB, with only a 1.5\% drop in NDS and 1.9\% drop in mAP compared with the full setting (keeping ratio of 1.0). 
This configuration is well-suited for edge devices or latency-critical applications. Likewise, using a keeping ratio of 0.50 or 0.75 provides a balanced trade-off between accuracy and resource consumption, making them more appropriate for deployment scenarios where the model needs to be integrated with downstream tasks.

\subsubsection{Generalization Ability of Sparse Selector}
To further evaluate the versatility of Sparse Selector, we integrate it into several representative paradigms, including multi-modality, LiDAR-only, and camera-only temporal methods. 
For the multi-modality setup~\cite{yan2023cross}, the input image resolution is set to $320 \times 800$ pixels with voxel dimensions of $0.1 \times 0.1$ meters. As shown in Tab.~\ref{tab:ge}, incorporating Sparse Selector improves the baseline by +0.7\% mAP and +0.8\% NDS under the same configuration, while reducing 105.5 GFLOPs and achieving a +1.4 FPS speedup. For the LiDAR-only paradigm, we adopt TransFusion (LiDAR-only) as the baseline, where adding Sparse Selector yields a +0.4\% NDS improvement and reduces 78.6 GFLOPs. Moreover, we also compare with the temporal camera-only baseline StreamPETR~\cite{Wang_2023_ICCV}. 
Following its original configuration, we set the sliding window to eight frames. 
With this setting, Sparse Selector reduces 86.2~GFLOPs while improving NDS by $+0.4\%$. These results collectively verify the plug-and-play capability of Sparse Selector, demonstrating its ability to reduce computational overhead while boosting performance across diverse modalities.

% Additionally, we conduct an ablation study to investigate the interaction between the LiDAR and image keeping ratios, as shown in Fig.~\ref{fig:twosubfigTwo}. We find that when only 25\% of the image tokens are retained, increasing the keeping ratio of LiDAR tokens leads to at most a 0.2\% improvement in mAP. However, when the keeping of image tokens increases to 50\%, mAP surprisingly improves by 1.7\%, demonstrating the significant contribution of sufficient image foreground information to model performance. Furthermore, when the keeping ratio of both LiDAR and image tokens exceeds 50\%, there is no significant impact on mAP, indicating that our sampling algorithm successfully captures the critical foreground information necessary for the prediction head.

\begin{table}[htbp]
\centering
\caption{
Ablation study of loss functions for distribution supervision. CE denotes cross-entropy loss, FL denotes focal loss~\cite{lin2017focal}, and CBS denotes class-balanced supervision loss. $\lambda$ and $\rho$ indicate the adjustment weight and the keeping ratio, respectively; in our experiments, $\rho$ is fixed at $0.5$. ``T\text{-}Cone'' denotes Traffic Cone. Metrics for Barrier and T\text{-}Cone are evaluated with a $0.5\,\text{m}$ threshold.
}

\label{tab:topk}
\resizebox{\linewidth}{!}{\begin{tabular}{@{}c|cc|cccc@{}}
\toprule
Loss& $\lambda$ & $\rho$ & NDS$\uparrow$ & mAP$\uparrow$ & AP (Barrier)$\uparrow$& AP (T-Cone)$\uparrow$ \\
\midrule
 CE& -&0.5 & 59.5 & 57.6 & 61.4 &  60.4\\
 FL& -&0.5 & 67.3 & 61.5 & 64.7& 63.8 \\
\midrule
  \multirow{5}{*}{CBS} & 1.0& 0.5 & 71.3 &69.2 & 74.8 & 73.5\\
  &  1.5 &  0.5 &  \textbf{71.9} &  \textbf{70.0} &  \textbf{75.6} &  \textbf{74.3}\\
  & 2.0& 0.5 & \textbf{71.9} & \textbf{70.0} & \textbf{75.6} & 74.2\\   
 % \rowcolor{gray!15} 
 & 2.5& 0.5 & 71.4 & 69.6 & \textbf{75.6} & 73.8\\
  & 3.0 & 0.5 & 70.3 & 68.3 & 73.2& 71.7\\
 % \checkmark&\checkmark & \checkmark&- & - \\
\bottomrule
\end{tabular}}
\end{table}
% \begin{table}[t]
% \centering
% \scriptsize

% \setlength{\tabcolsep}{4pt}
% \renewcommand{\arraystretch}{1.2}
% \begin{tabular}{c|c|c|c|c|c|c|c|c}
% \toprule
% Sinu & MAPE & Im & PC & NDS$\uparrow$ & mAP$\uparrow$ & mATE$\downarrow$ & mASE$\downarrow$ & mAOE$\downarrow$ \\
% \midrule
% \checkmark &  &  &  & 70.3 & 68.7 & 28.4 & 24.6 & 30.4 \\
%  % & \checkmark &  &  & 57.4 & 54.2 & 53.9 & 27.1 & 43.3 \\
%  % &  & \checkmark &  & 67.4 & 64.7 & 36.1 & 24.5 & 34.2 \\
% \rowcolor{gray!10}
%  & \checkmark & \checkmark & \checkmark & \textbf{72.4} & \textbf{70.0} & \textbf{27.9} & \textbf{23.5} & 30.8 \\
% \bottomrule
% \end{tabular}
% \end{table}

\begin{table}[t]
\centering
\scriptsize
\caption{
Analysis of different positional encodings and different sampling points $d$ for our proposed Ray PE. ``Sine'' denotes the vanilla sinusoidal positional encoding~\cite{vaswani2017attention}, while ``Learnable'' refers to gradient-updated positional embeddings. 
}
\label{tab:raype}
\resizebox{\linewidth}{!}{
\begin{tabular}{c|c|ccc}
\toprule
points $d$ & spatial pos. & NDS$\uparrow$ & mAP$\uparrow$ & mASE$\downarrow$\\
\midrule
-  & Learnable & 69.6 & 67.5 & 25.6 \\
-  & Sine      & 70.3 & 68.7 & 23.9 \\
\midrule
8  & Ray PE    & 72.2 & 69.8 & 23.9 \\
\rowcolor{gray!10}
16 & Ray PE    & \textbf{72.4} & \textbf{70.0} & \textbf{23.5} \\
20 & Ray PE    & {72.3} & \textbf{70.0} & \textbf{23.5} \\
24 & Ray PE    & 72.0 & \textbf{70.0} & 24.4 \\
\bottomrule
\end{tabular}
}
\end{table}

\subsection{Strong Robustness}
To validate the robustness of our method, following the robustness benchmark \cite{dong2023benchmarking, yan2023cross}, we evaluate our method under various harsh environments, including four challenging conditions: (1) missing LiDAR data in the range field ($-90^\circ$, $90^\circ$), (2) missing the entire LiDAR sensor, (3) missing the most critical front camera, and (4) missing all cameras. As shown in Tab.~\ref{tab:sensor_malfunction}, the results demonstrate that our sparse detector exhibits strong resilience to sensor malfunctions.
This is because our multi-modal method does not rely heavily on any single modality. Notably, when camera information is missing, our method achieves 70.6\% mAP, still performing close to SoTA, whereas TransFusion fails when LiDAR is absent due to its two-stage design. SparseFusion, requiring additional detection heads, cannot function when either the camera or LiDAR is completely unavailable.

\subsection{Failure Cases and Limitations} 
\begin{figure*}[thbp]
    \centering
    \includegraphics[width=\textwidth]{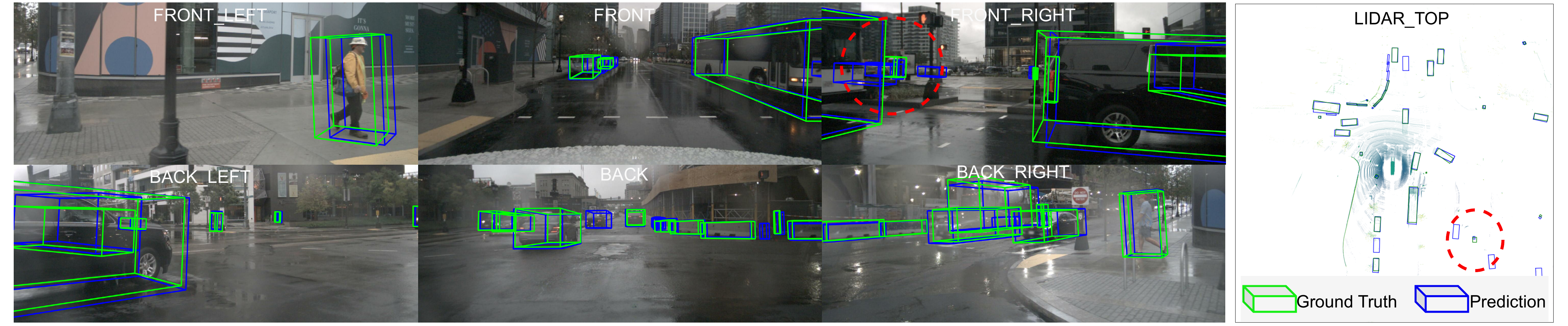}
    \caption{
    Visualization results of CrossRay3D. On the BEV plane (right), ground truth and predictions are shown in green and blue rectangles, respectively, while failure cases are highlighted with red circles. The keeping ratio is fixed at $0.25$.
}
    \label{fig:hard}
\end{figure*}

We present detection results under challenging weather conditions in Fig.~\ref{fig:hard}. To validate the performance of our algorithm, the keeping ratio is set to 0.25. CrossRay3D achieves impressive results on crowded objects within a detection range of 30\,m. However, our method still produces orientation errors for small objects at farther distances. Under foggy or rainy conditions, such orientation errors on distant targets are relatively tolerable and acceptable. For real-world deployment, although CrossRay3D incorporates a token selection mechanism for both LiDAR and camera data to reduce the resource consumption of the decoder, the computational efficiency of the backbone still requires further optimization.

\section{Conclusion}
We explore the key challenges faced by sparse detectors and propose CrossRay3D, an end-to-end sparse multimodal detector that achieves comparable accuracy while significantly reducing computational consumption. The core component of CrossRay3D is a multimodal token discrimination strategy, which considers both geometry and distribution to achieve optimal token selection. Additionally, we introduce Ray PE to facilitate the spatial alignment of multimodal tokens while mitigating distribution discrepancies across modalities. Experimental results demonstrate that CrossRay3D has become the SOTA method for token pruning in multimodal models. Our work provides valuable insights into the design of sparse detectors.

\begin{table}[t]
\centering
\scriptsize
\caption{{Comparisons on the Argoverse 2 validation set. We evaluate across 26 object categories within a range of 150 meters. C-Cone: construction cone. Some categories are excluded from the table due to the limited number of instances they contain. However, the average results consider all categories, even those that are omitted. Following PolFusion~\cite{deng2024poifusion}, the voxel size of our CrossRay3D is (0.2, 0.2, 0.2), the image backbone is ResNet-50, and the image resolution is 960×640.}}
\label{tab:av2}
\setlength{\tabcolsep}{5pt}
\renewcommand{\arraystretch}{1.2}
\begin{tabular}{l|l|ccccc}
\toprule
 & Methods & Average & Vehicle & Pedestrian & C-Cone & Bicycle \\
\midrule
\multirow{6}{*}{mAP}
 & CenterPoint~\cite{yin2021center}       & 22.0 & 67.6 & 46.5 & 29.5 & 24.5 \\
 & Far3D~\cite{jiang2024far3d}            & 24.4 & -- & -- & -- & -- \\
 & FSF~\cite{li2024fully}                 & 33.2 & 70.8 & 60.8 & 51.7 & 38.6 \\
 & CMT~\cite{yan2023cross}                & 36.1 & 71.9 & 61.2 & 59.5 & 40.3 \\
 & PolFusion~\cite{deng2024poifusion}     & 40.6 & 77.6 & 70.6 & \textbf{64.6} & 55.1 \\
 & \textbf{CrossRay3D (ours)}                             & \textbf{42.4} & \textbf{79.1} & \textbf{72.9} & 64.5 & \textbf{57.0} \\
\midrule
\multirow{6}{*}{CDS}
 & CenterPoint~\cite{yin2021center}       & 17.6 & 57.2 & 35.7 & 22.4 &  19.6 \\ 
 & Far3D~\cite{jiang2024far3d}            & 18.1 & -- & -- & -- & -- \\
 & FSF~\cite{li2024fully}                 & 25.5 & 59.6 & 48.5 & 37.3 &  32.0 \\
 & CMT~\cite{yan2023cross}                & 27.8 & 62.2 & 46.8 & 42.5 & 29.8 \\
 & PolFusion~\cite{deng2024poifusion}     & 31.6 & 66.5 & 54.8 & \textbf{47.8} & 42.8 \\
 & \textbf{CrossRay3D (ours)}                             & \textbf{33.1} & \textbf{67.0} & \textbf{57.2} & 46.7 & \textbf{43.4} \\
\bottomrule
\end{tabular}
\end{table}

\bibliographystyle{IEEEtran}
\bibliography{mybib}

\end{document}